\documentclass{article}

\usepackage{microtype}
\usepackage{graphicx}
\usepackage{subcaption}
\usepackage{booktabs}
\usepackage{hyperref}

\usepackage[accepted]{icml2026}

\usepackage[utf8]{inputenc}
\usepackage[T1]{fontenc}
\usepackage{amsmath}
\usepackage{amssymb}
\usepackage{amsfonts}
\usepackage{mathtools}
\usepackage{amsthm}
\usepackage{nicefrac}
\usepackage{xcolor}
\usepackage{multirow}
\usepackage{array}
\usepackage{tikz}
\usetikzlibrary{arrows.meta,positioning,fit,calc}
\usepackage{algorithm}
\usepackage{algorithmic}

\usepackage[capitalize,noabbrev]{cleveref}

\theoremstyle{plain}

\theoremstyle{definition}

\theoremstyle{remark}

\newcommand{\method}{RETROSPECT}
\newcommand{\genmodel}{ChemAlign Transformer}
\newcommand{\uspto}{USPTO-50K}
\definecolor{okblue}{HTML}{0072B2}
\definecolor{okorange}{HTML}{E69F00}
\definecolor{okgreen}{HTML}{009E73}
\definecolor{okred}{HTML}{D55E00}
\definecolor{okpurple}{HTML}{CC79A7}

\icmltitlerunning{RETROSPECT: Proposal and Reranking for Single-Step Retrosynthesis}

\begin{document}

\twocolumn[
  \icmltitle{\method: RETROsynthesis via Sequential Prediction,\\
  and Chemically Transformed-ranking}

  \icmlsetsymbol{equal}{*}

  \begin{icmlauthorlist}
    \icmlauthor{Raja Sekhar Pappala}{equal,mstack}
    \icmlauthor{Shreyas Vinaya Sathyanarayana}{equal,mstack}
    \icmlauthor{Ronit Kumar Choudhary}{equal,mstack}
    \icmlauthor{Arjun Verma}{equal,mstack}
    \icmlauthor{Deepak Warrier}{equal,mstack}
  \end{icmlauthorlist}

  \icmlaffiliation{mstack}{Mstack AI}
  \icmlcorrespondingauthor{Shreyas Vinaya Sathyanarayana}{shreyas.v@mstack.co}

  \vskip 0.3in
]

\printAffiliationsAndNotice{\textsuperscript{*}Equal contribution.}

\begin{abstract}
Single-step retrosynthesis needs both accurate first-ranked suggestions and candidate lists that are rich enough for downstream selection. We study this as a proposal-selection decomposition. Our system, \method, combines a single Transformer proposal model, which we call the \genmodel, with a LambdaMART reranker over structural, reaction-template, upstream-score, and optional DFT-derived descriptors. The generator is trained with hybrid root-aligned and random SMILES augmentation, Pre-LayerNorm, tied embeddings, exponential moving average weights, and a differentiable atom-balance auxiliary loss. On the full \uspto\ test set of 5,007 reactions, the generator reaches 55.00\% top-1 and 86.18\% top-10 exact-match accuracy with 99.86\% top-1 validity. On the merged candidate-pool benchmark used for reranking, which contains 5,007 test products and about 111 candidates per product, a LambdaMART model trained on the structural feature set reaches 59.4\% top-1 with 0.7171 mean reciprocal rank. Feature ablations show that upstream proposal score and template-frequency statistics provide most of the reranking signal, while DFT and reaction-center DFT features provide smaller and less consistent gains. These results support a modular view of retrosynthesis: stronger single-model proposal and learned candidate selection are complementary, and the proposal model can serve as a drop-in component for ensemble systems such as RetroChimera \citep{maziarz2024retrochimera}.
\end{abstract}

\section{Introduction}
\label{sec:intro}

Retrosynthesis asks which precursor molecules can produce a target molecule. The problem is central to computer-aided synthesis planning, where a single-step model is repeatedly called inside a multi-step search procedure \citep{corey1969computer,coley2018machine,segler2018planning}. A useful single-step model must therefore satisfy two requirements. It must place a correct disconnection near the top of its ranked list, and it must preserve enough plausible alternatives for a planner or chemist to recover when the first suggestion is unavailable, unsafe, or strategically poor.

Many recent retrosynthesis systems fold proposal and ranking into one stage. Template-based methods classify or retrieve reaction templates \citep{dai2019retrosynthesis,somnath2021learning,chen2021deep,gainski2024retrogfn}. Template-free methods generate precursor strings or graphs directly \citep{liu2017retrosynthetic,tetko2020state,wan2022retroformer,maziarz2024retrochimera,igashov2024retrobridge,yadav2025retrosynflow,han2024editretro}. Both families ultimately produce ranked candidates, but the mechanisms that enumerate plausible disconnections and the mechanisms that decide their order are not necessarily the same.

This paper focuses on that separation. We develop a stronger single proposal model, the \genmodel, then study a learning-to-rank stage that reranks merged candidate pools produced by beam search and SMILES augmentation. The proposal model is intentionally single-model rather than ensemble-based. This lets us ask two cleaner questions. First, how much can a carefully trained Transformer proposal model achieve on its own? Second, once a proposal pool exists, which feature families actually improve its ordering?

Our contributions are:
\begin{enumerate}
    \item We present \method, a modular proposal-plus-reranking framework for single-step retrosynthesis that cleanly separates candidate generation from candidate selection.
    \item We develop the \genmodel, a stronger Augmented Transformer variant using hybrid root-aligned/random SMILES augmentation, Pre-LayerNorm, exponential moving average weights, tied token embeddings, and a differentiable atom-balance auxiliary loss.
    \item We implement a LambdaMART reranker over upstream score, structural descriptors, reaction-template descriptors, and optional DFT-derived features, and document the train-split freezing needed for frequency-style statistics.
    \item We provide ablations that show where the reranking signal comes from: proposal score and template-derived features matter most in the current setup, while DFT and reaction-center DFT features are weaker and should be treated cautiously.
\end{enumerate}

The resulting picture is deliberately conservative. Our verified numbers support a strong single-model proposal system and a useful reranking study, not a claim that every component uniformly surpasses the best published end-to-end baselines. Instead, we argue that the proposal model is competitive as a standalone system and attractive as a drop-in candidate source for ensemble frameworks such as RetroChimera \citep{maziarz2024retrochimera}.

\section{Related work}
\label{sec:related}

\paragraph{Template-based retrosynthesis.}
Template-based systems classify or retrieve reaction templates, then apply them to a product molecule. GLN models product-template and reactant-template compatibility with graph logic \citep{dai2019retrosynthesis}. GraphRetro decomposes prediction into reaction-center identification and leaving-group completion \citep{somnath2021learning}. LocalRetro predicts local reactivity with global attention and remains a strong high-$k$ template-based baseline \citep{chen2021deep}. RetroGFN uses generative flow networks to diversify template selections \citep{gainski2024retrogfn}. These methods are interpretable and chemically constrained, but their proposal mechanism is still tied to a template inventory.

\paragraph{Template-free and semi-template methods.}
Sequence-to-sequence retrosynthesis treats product-to-reactant prediction as translation \citep{liu2017retrosynthetic,zheng2019predicting,tetko2020state}. Later systems add graph structure, edit operations, and reaction-center priors, including G2Gs \citep{shi2020graph}, MEGAN \citep{sacha2021molecule}, GTA \citep{seo2021gta}, Graph2SMILES \citep{tu2022permutation}, Retroformer \citep{wan2022retroformer}, Graph2Edits \citep{zhong2023graph2edits}, G2Retro \citep{chen2023g2retro}, and EditRetro \citep{han2024editretro}. Recent frontier systems include RetroChimera, which ensembles complementary proposal models with learned molecule-space ranking \citep{maziarz2024retrochimera}, and Retro SynFlow, which applies discrete flow matching from synthons to reactants \citep{yadav2025retrosynflow}. Our work is closest in spirit to systems that explicitly acknowledge multiple proposal hypotheses, but we study the proposal model and reranker as separate units rather than introducing a new ensemble.

\paragraph{Aligned reaction representations.}
Root-aligned SMILES reduce source-target edit distance by starting product and reactant SMILES from corresponding atoms \citep{zhong2022root}. This representation is especially helpful for reactions because much of the molecular graph is preserved. Our generator uses a hybrid augmentation strategy, mixing root-aligned and random SMILES, so the model receives both tightly aligned supervision and traversal diversity.

\paragraph{Learning to rank for retrosynthesis.}
Ranking appears implicitly in beam search, template probabilities, and ensemble aggregation. RetroChimera uses learned ranking to combine predictions from several complementary models \citep{maziarz2024retrochimera}. Retro-Rank-In formulates inorganic retrosynthesis as ranking precursor candidates in a shared latent space \citep{prein2025retro}. We focus on organic USPTO-50K retrosynthesis and study how a listwise LambdaMART reranker interacts with a single strong proposal model.

\section{Method}
\label{sec:method}

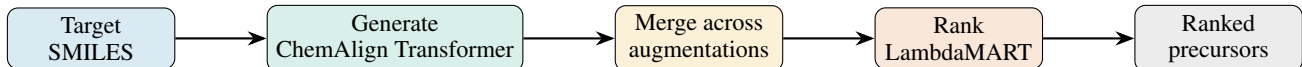
\begin{figure*}[t]
\centering
\resizebox{\linewidth}{!}{%
\begin{tikzpicture}[
  node distance=1.2cm,
  box/.style={rectangle, draw, rounded corners, align=center, minimum height=0.8cm, minimum width=2.2cm, font=\small},
  arrow/.style={-{Stealth[length=2.5mm]}, thick}
]
\node[box, fill=okblue!15] (target) {Target\\SMILES};
\node[box, fill=okgreen!15, right=of target] (gen) {Generate\\\genmodel};
\node[box, fill=okorange!15, right=of gen] (merge) {Merge across\\augmentations};
\node[box, fill=okred!15, right=of merge] (rank) {Rank\\LambdaMART};
\node[box, fill=gray!15, right=of rank] (out) {Ranked\\precursors};
\draw[arrow] (target) -- (gen);
\draw[arrow] (gen) -- (merge);
\draw[arrow] (merge) -- (rank);
\draw[arrow] (rank) -- (out);
\end{tikzpicture}%
}
\caption{\method\ separates candidate proposal from candidate selection. The generator produces candidates under multiple SMILES traversals, these candidates are merged and deduplicated, and a listwise reranker reorders the resulting proposal pool.}
\label{fig:pipeline}
\end{figure*}

Given target molecule $T$, \method\ returns a ranked list of precursor sets $P_1,\ldots,P_K$. \Cref{fig:pipeline} shows the pipeline.

\subsection{Proposal model: the \genmodel}

The generator is an encoder-decoder Transformer over tokenized SMILES. It builds on the Augmented Transformer \citep{tetko2020state}, but we rename the model \genmodel\ to emphasize reaction-aligned supervision, stronger optimization, and chemical regularization. The model uses six encoder layers, six decoder layers, hidden size 512, eight attention heads, feed-forward size 2048, dropout 0.1, sinusoidal positional encodings up to length 1000, and an 81-token vocabulary built from the training split only using a Molecular Transformer-style regex tokenizer \citep{schwaller2019molecular}. Source sequences append \texttt{<EOS>}; target sequences use \texttt{<BOS>} and \texttt{<EOS>} delimiters.

Training uses 20-fold offline augmentation over 40,008 training reactions, yielding 800,160 source-target pairs. Sixteen augmented pairs per reaction use root-aligned SMILES, which align product and reactant traversals around corresponding atoms \citep{zhong2022root}, and four use random SMILES to preserve traversal invariance. Randomly augmented multi-fragment reactants are canonically fragment-sorted to reduce output-order variance, while root-aligned reactants retain alignment order. Validation and test inputs remain canonical. We use Pre-LayerNorm for stable optimization \citep{xiong2020layer}, three-way weight tying among encoder embeddings, decoder embeddings, and output projection \citep{press2017using}, Xavier-style initialization, and EMA weights with decay 0.999 for validation and inference.

Optimization details are summarized in \Cref{app:gen_config}. We train with token-normalized cross-entropy, no label smoothing, Adam with $\beta_1=0.9$, $\beta_2=0.998$, and $\epsilon=10^{-9}$, and a Noam schedule with factor 2.0 and 8,000 warmup steps. Batches are packed by target length up to 16,384 tokens per micro-batch, accumulated for two steps, and trained with AMP. Validation runs every 2,000 optimizer steps, and the best EMA checkpoint occurs at step 20,000.

The auxiliary atom-balance loss penalizes deviations between expected element counts under the decoder softmax and ground-truth element fractions. Let $z_t$ be decoder logits at position $t$ and let $A$ map each token to element counts for 12 elements. We compute expected counts $\hat a = \sum_t \mathrm{softmax}(z_t) A$ and add an L1 penalty with coefficient $0.1$. This loss is not a hard validity constraint, but it biases the model away from mass-balance violations while preserving differentiability.

\subsection{LambdaMART reranking over merged candidate pools}

The reranker is an XGBoost LambdaMART model trained with the listwise \texttt{rank:ndcg} objective \citep{chen2016xgboost,burges2010ranknet}. Candidate pools are produced by running the generator under multiple product traversals, then merging and deduplicating the resulting hypotheses. For each target-candidate pair, we compute four feature blocks: structural descriptors, optional DFT-derived descriptors, reaction-template descriptors, and the upstream proposal score. Structural descriptors include pharmacophore counts, functional-group indicators, atom-count deltas, and Morgan/MACCS fingerprint similarities \citep{rogers2010extended}. Reaction-template descriptors include whether a template can be extracted for the pair, hashed template identifiers at multiple radii, and train-frozen template-frequency statistics. DFT descriptors, when available, include reaction-level differences in HOMO, LUMO, gap, dipole, hardness, softness, and derived frontier-orbital features.

Before ranking, candidates are canonicalized to a shared precursor-set representation and merged by canonicalized precursor SMILES. If the same precursor set is proposed multiple times, we keep one merged row, retain source provenance, and carry forward the best available upstream score for that precursor set. LambdaMART is trained groupwise within each product, never across products. The training targets use graded relevance labels so exact precursor-set matches receive the highest gain while partial fragment overlap receives weaker positive labels. In the current tuned models, the upstream score is the single strongest feature, template-frequency features provide the next-largest gains, and DFT features remain secondary. Additional protocol details and the ranker hyperparameters are summarized in \Cref{app:ranker_protocol,app:ranker_tuning}.

\section{Experiments}
\label{sec:experiments}

\paragraph{Dataset.}
We evaluate on \uspto, using the standard split of 40,008 training, 5,000 validation, and 5,007 test reactions \citep{schneider2016whats,liu2017retrosynthetic}. We report exact-match top-$k$ accuracy after canonicalizing predicted and ground-truth precursor sets. Unless noted otherwise, reaction class is unknown at test time.

\paragraph{Preprocessing and augmentation.}
Raw USPTO reactions follow the atom-mapped format popularized by Lowe \citep{lowe2012extraction}. We discard reagents, remove atom mapping, and canonicalize products and precursor sets with RDKit before training. The generator uses a double-canonicalization pass for stereochemical consistency, then applies 20 offline augmentations per training reaction, while validation and test examples remain canonical. Multi-fragment ordering is a real modeling issue in this benchmark: 70.7\% of training reactions have two precursor fragments, 29.1\% have one, and 0.2\% have three. Randomly augmented reactant fragments are therefore canonically sorted during preprocessing, while R-SMILES variants preserve product-aligned fragment order.

\paragraph{Proposal and reranking protocols.}
The full-test proposal result uses canonical-input inference only and is the fairest single-model comparison to prior seq2seq systems. The reranking study uses larger merged candidate pools: train and validation pools are built from fewer beam-search augmentations per product than the test pool, while the test reranking benchmark contains 5,007 products with about 111 candidates per product on average. This mismatch makes the reranker closer to a robust reordering model than a memorizer of one fixed candidate width. We freeze all frequency-style statistics on the training split before applying them to validation and test examples.

\paragraph{Baselines.}
Our literature survey includes classical template-based baselines, modern template-free baselines, and recent strong systems including EditRetro \citep{han2024editretro}, RetroChimera \citep{maziarz2024retrochimera}, and Retro SynFlow \citep{yadav2025retrosynflow}. We compare the \genmodel\ directly against published single-step systems in the main table. We report the reranking ablations separately because they operate on a merged candidate-pool benchmark rather than the full 5,007-reaction end-to-end test set.

\subsection{Single-model proposal results}

\Cref{tab:main} reports the generator alone. We position RETROSPECT as a single-model system rather than an ensemble, and this table should be read as the main evidence for that claim. The best full-test RETROSPECT generator checkpoint, which is also our best single-model configuration, reaches 55.00\% top-1 and 86.18\% top-10 with 99.86\% top-1 validity. A 15K-reaction ablation shows that hybrid root-aligned SMILES is the dominant design choice, improving top-1 by 9.23 percentage points over the scaled baseline. Pre-LayerNorm, EMA, and atom-balance regularization add another 1.54 points. The cumulative ablation table appears in \Cref{app:generator_ablation}.

\begin{table*}[t]
\centering
\caption{Top-$k$ exact-match accuracy on the \uspto\ test set (5,007 reactions) with reaction class unknown. The \genmodel\ row reports beam search alone; the \method\ row adds a LambdaMART reranker trained on the structural feature set over the merged candidate pool of about 111 candidates per product. TB = template-based, ST = semi-template, TF = template-free.}
\label{tab:main}
\small
\begin{tabular}{llcccc}
\toprule
Type & Method & Top-1 & Top-3 & Top-5 & Top-10 \\
\midrule
TB & GLN \citep{dai2019retrosynthesis} & 52.5 & 74.7 & 81.2 & 87.9 \\
TB & LocalRetro \citep{chen2021deep} & 53.4 & 77.5 & 85.9 & 92.4 \\
ST & GraphRetro \citep{somnath2021learning} & 53.7 & 68.3 & 72.2 & 75.5 \\
ST & G2Retro \citep{chen2023g2retro} & 54.1 & 74.1 & 81.2 & 86.7 \\
TF & Graph2Edits \citep{zhong2023graph2edits} & 55.1 & 77.3 & 83.4 & 89.4 \\
TF & R-SMILES \citep{zhong2022root} & 56.3 & 79.2 & 86.2 & 91.0 \\
TF & RetroChimera \citep{maziarz2024retrochimera} & 59.6 & \textbf{82.8} & \textbf{89.2} & \textbf{94.2} \\
TF & Retro SynFlow \citep{yadav2025retrosynflow} & $60.0_{\pm 0.22}$ & $77.9_{\pm 0.13}$ & $82.7_{\pm 0.15}$ & $85.3_{\pm 0.19}$ \\
TF & EditRetro \citep{han2024editretro} & \textbf{60.8} & 80.6 & 86.0 & 90.3 \\
\midrule
Ours & \genmodel\ only & 55.00 & 76.13 & 81.33 & 86.18 \\
Ours & \method, structural & 59.4 & \textbf{82.02} & \textbf{87.51} & \textbf{93.06} \\
\bottomrule
\end{tabular}
\end{table*}

\subsection{Reranking on merged candidate pools}

The full \method\ row in \Cref{tab:main} adds a LambdaMART reranker on the merged candidate pool of about 111 candidates per product. Reranking lifts top-1 from 55.00\% to 59.4\%, top-10 from 86.18\% to 93.06\%, and reaches 0.7171 mean reciprocal rank. Earlier feature ablations over a V1 proposal pool, reported in \Cref{app:ranker_variants}, show that the choice of feature set shifts top-$k$ by less than one percentage point and that adding DFT or reaction-center DFT does not clearly dominate simpler sets, so we run the V2 reranker on the structural feature set only. Full protocol and tuning details appear in \Cref{app:ranker_protocol,app:ranker_tuning}.

\section{Analysis}
\label{sec:analysis}

\paragraph{A stronger proposal model does not eliminate the value of selection.}
The generator is a competitive single model, but the reranking study shows that proposal and ordering still capture different signals. The proposal model determines whether plausible candidates enter the pool. The reranker exploits upstream score, template-derived priors, and structural compatibility to reorder those candidates once they exist.

\paragraph{Upstream score and template priors dominate the current reranker.}
The feature-importance profile and ablations indicate that the proposal score is the strongest single feature, while template-frequency and template-identity features add a large fraction of the remaining gain. This is useful scientifically because it narrows where future effort should go: better candidate scoring and better train-split template statistics are currently more valuable than larger DFT stacks.

\paragraph{DFT and reaction-center DFT are promising but not yet central.}
The DFT feature families can improve certain higher-$k$ metrics or MRR, especially in reaction-center variants, but the gains are small and inconsistent across settings. In the current paper, they should therefore be treated as exploratory chemistry-aware features rather than the main explanation for performance.

\paragraph{The proposal model is modular enough for ensemble systems.}
RetroChimera already shows that complementary proposal mechanisms and learned ranking can outperform any one component \citep{maziarz2024retrochimera}. Our results suggest a practical use case: the \genmodel\ is a stronger single-model proposal module than earlier Augmented Transformer baselines, so it is a natural component to test inside ensemble-and-ranking frameworks rather than only as a standalone decoder.

\section{Limitations and broader impact}
\label{sec:limitations}

The main limitation is that results are reported on \uspto, a widely used but small and biased benchmark extracted from patent reactions. The proposal result and the reranking result answer related but different questions: the full-test table measures standalone single-model exact-match accuracy, while the reranking study measures ordering quality on merged candidate pools with 5,007 products and about 111 candidates per product. If the correct precursor is absent from the proposal pool, no reranker can recover it. DFT-derived features and reaction-center features are also under-validated relative to the generator, and their benefits remain modest in the current ablations. Finally, exact-match accuracy rewards reproducing the recorded patent precursor set, even when other chemically valid disconnections exist.

The positive impact of improved retrosynthesis is faster synthesis planning for drug discovery, materials design, and chemical supply chains. Potential risks include accelerating access to harmful compounds or over-trusting algorithmic routes without expert review. We view \method\ as a decision-support system for trained chemists rather than an autonomous synthesis planner. Practical deployment should include building-block availability checks, condition prediction, safety filters, and human review.

\section{Conclusion}
\label{sec:conclusion}

We presented \method, a proposal-plus-reranking framework for single-step retrosynthesis. The \genmodel\ provides a strong single-model proposal system through aligned SMILES supervision and improved optimization, while LambdaMART shows that learned candidate selection still adds useful signal once a rich proposal pool is available. The cleanest empirical conclusions are threefold. First, root-aligned augmentation is the dominant contributor to proposal quality. Second, reranking gains in the current setup come mainly from proposal score and template-derived priors. Third, DFT-based features remain exploratory. This decomposition makes the system scientifically easier to analyze and practically easier to reuse: the proposal model can stand alone, and it can also be inserted into ensemble systems such as RetroChimera for future work.

\bibliography{references}

@inproceedings{dai2019retrosynthesis,
  title={Retrosynthesis Prediction with Conditional Graph Logic Network},
  author={Dai, Hanjun and Li, Chengtao and Coley, Connor W and Dai, Bo and Song, Le},
  booktitle={Advances in Neural Information Processing Systems (NeurIPS)},
  volume={32},
  year={2019}
}

@inproceedings{somnath2021learning,
  title={Learning Graph Models for Retrosynthesis Prediction},
  author={Somnath, Vignesh Ram and Bunne, Charlotte and Coley, Connor W and Krause, Andreas and Barzilay, Regina},
  booktitle={Advances in Neural Information Processing Systems (NeurIPS)},
  volume={34},
  year={2021}
}

@article{chen2021deep,
  title={Deep Retrosynthetic Reaction Prediction using Local Reactivity and Global Attention},
  author={Chen, Shuan and Jung, Yousung},
  journal={JACS Au},
  volume={1},
  number={10},
  pages={1612--1620},
  year={2021},
  publisher={ACS Publications}
}

@article{gainski2024retrogfn,
  title={RetroGFN: Diverse and Feasible Retrosynthesis using GFlowNets},
  author={Gai{\'n}ski, Piotr and Koziarski, Micha{\l} and Maziarz, Krzysztof and Segler, Marwin and Tabor, Jacek and {\'S}mieja, Marek},
  journal={arXiv preprint arXiv:2406.18739},
  year={2024}
}

@article{zheng2019predicting,
  title={Predicting Retrosynthetic Reactions Using Self-Corrected Transformer Neural Networks},
  author={Zheng, Shuangjia and Rao, Jiahua and Zhang, Zhongyue and Xu, Jun and Yang, Yuedong},
  journal={Journal of Chemical Information and Modeling},
  volume={60},
  number={1},
  pages={47--55},
  year={2020},
  publisher={ACS Publications}
}

@inproceedings{shi2020graph,
  title={A Graph to Graphs Framework for Retrosynthesis Prediction},
  author={Shi, Chence and Xu, Minkai and Guo, Hongyu and Zhang, Ming and Tang, Jian},
  booktitle={International Conference on Machine Learning (ICML)},
  pages={8818--8827},
  year={2020},
  organization={PMLR}
}

@article{tetko2020state,
  title={State-of-the-Art Augmented {NLP} Transformer Models for Direct and Single-Step Retrosynthesis},
  author={Tetko, Igor V and Karpov, Pavel and Van Deursen, Ruud and Godin, Guillaume},
  journal={Nature Communications},
  volume={11},
  number={1},
  pages={5575},
  year={2020},
  publisher={Nature Publishing Group}
}

@article{sacha2021molecule,
  title={Molecule Edit Graph Attention Network: Modeling Chemical Reactions as Sequences of Graph Edits},
  author={Sacha, Miko{\l}aj and B{\l}a{\.z}, Miko{\l}aj and Byrski, Piotr and W{\l}odarczyk-Pruszy{\'n}ski, Pawe{\l} and Jastrz{\k{e}}bski, Stanis{\l}aw},
  journal={Journal of Chemical Information and Modeling},
  volume={61},
  number={7},
  pages={3273--3284},
  year={2021},
  publisher={ACS Publications}
}

@inproceedings{seo2021gta,
  title={{GTA}: Graph Truncated Attention for Retrosynthesis},
  author={Seo, Seung-Woo and Song, You Young and Yang, June Yong and Bae, Seohui and Lee, Hankook and Shin, Jinwoo and Hwang, Sung Ju and Yang, Eunho},
  booktitle={Proceedings of the AAAI Conference on Artificial Intelligence},
  volume={35},
  pages={531--539},
  year={2021},
  doi={10.1609/aaai.v35i1.16131}
}

@article{tu2022permutation,
  title={Permutation Invariant Graph-to-Sequence Model for Template-Free Retrosynthesis and Reaction Prediction},
  author={Tu, Zhengkai and Coley, Connor W},
  journal={Journal of Chemical Information and Modeling},
  volume={62},
  number={15},
  pages={3503--3513},
  year={2022},
  publisher={ACS Publications}
}

@inproceedings{wan2022retroformer,
  title={Retroformer: Pushing the Limits of End-to-end Retrosynthesis Transformer},
  author={Wan, Yue and Hsieh, Chang-Yu and Liao, Ben and Jia, Shengyu},
  booktitle={International Conference on Machine Learning (ICML)},
  pages={22475--22490},
  year={2022},
  organization={PMLR}
}

@article{maziarz2024retrochimera,
  title={Chemist-Aligned Retrosynthesis by Ensembling Diverse Inductive Bias Models},
  author={Maziarz, Krzysztof and Liu, Guoqing and Misztela, Hubert and Tripp, Austin and Li, Jingxuan and Kornev, Alexey and Gai{\'n}ski, Piotr and Hoefling, Hanno and Fortunato, Mike and Gupta, Rianne and Segler, Marwin},
  journal={arXiv preprint arXiv:2412.05269},
  year={2024}
}

@inproceedings{igashov2024retrobridge,
  title={RetroBridge: Modeling Retrosynthesis with Markov Bridges},
  author={Igashov, Ilia and Schneuing, Arne and Segler, Marwin and Bronstein, Michael and Correia, Bruno},
  booktitle={International Conference on Learning Representations (ICLR)},
  year={2024}
}

@article{yadav2025retrosynflow,
  title={{RETRO} {S}yn{F}low: Discrete Flow Matching for Accurate and Diverse Single-Step Retrosynthesis},
  author={Yadav, Robin and Yan, Qi and Wolf, Guy and Bose, Avishek Joey},
  journal={arXiv preprint arXiv:2506.04439},
  year={2025}
}

@article{prein2025retro,
  title={Retro-Rank-In: A Ranking-Based Approach for Inorganic Materials Synthesis Planning},
  author={Prein, Thorben and Pan, Elton and Haddouti, Sami and Lorenz, Marco and Jehkul, Janik and Wilk, Tobiasz and Moran, Cormac and Fotiadis, Michail P and Toshev, Artur P and Olivetti, Elsa},
  journal={arXiv preprint arXiv:2502.04289},
  year={2025}
}

@article{schneider2016whats,
  title={What's What: The (Nearly) Definitive Guide to Reaction Role Assignment},
  author={Schneider, Nadine and Stiefl, Nikolaus and Landrum, Gregory A},
  journal={Journal of Chemical Information and Modeling},
  volume={56},
  number={12},
  pages={2336--2346},
  year={2016},
  publisher={ACS Publications},
  doi={10.1021/acs.jcim.6b00564}
}

@article{liu2017retrosynthetic,
  title={Retrosynthetic Reaction Prediction Using Neural Sequence-to-Sequence Models},
  author={Liu, Bowen and Ramsundar, Bharath and Kawthekar, Prasad and Shi, Jingbo and Gomes, Joseph and Nguyen, Quang Luu and Ho, Stephen and Sloane, Jack and Wender, Paul and Pande, Vijay},
  journal={ACS Central Science},
  volume={3},
  number={10},
  pages={1103--1113},
  year={2017},
  publisher={ACS Publications}
}

@inproceedings{chen2016xgboost,
  title={{XGBoost}: A Scalable Tree Boosting System},
  author={Chen, Tianqi and Guestrin, Carlos},
  booktitle={Proceedings of the 22nd ACM SIGKDD International Conference on Knowledge Discovery and Data Mining},
  pages={785--794},
  year={2016},
  organization={ACM},
  doi={10.1145/2939672.2939785}
}

@article{rogers2010extended,
  title={Extended-Connectivity Fingerprints},
  author={Rogers, David and Hahn, Mathew},
  journal={Journal of Chemical Information and Modeling},
  volume={50},
  number={5},
  pages={742--754},
  year={2010},
  publisher={ACS Publications}
}

@article{corey1969computer,
  title={Computer-Assisted Design of Complex Organic Syntheses},
  author={Corey, Elias J and Wipke, W Todd},
  journal={Science},
  volume={166},
  number={3902},
  pages={178--192},
  year={1969},
  publisher={American Association for the Advancement of Science}
}

@article{schwaller2019molecular,
  title={Molecular Transformer: A Model for Uncertainty-Calibrated Chemical Reaction Prediction},
  author={Schwaller, Philippe and Laino, Teo and Gaudin, Th{\'e}ophile and Bolgar, Peter and Hunter, Christopher A and Bekas, Costas and Lee, Alpha A},
  journal={ACS Central Science},
  volume={5},
  number={9},
  pages={1572--1583},
  year={2019},
  publisher={ACS Publications}
}

@article{coley2018machine,
  title={Machine Learning in Computer-Aided Synthesis Planning},
  author={Coley, Connor W and Green, William H and Jensen, Klavs F},
  journal={Accounts of Chemical Research},
  volume={51},
  number={5},
  pages={1281--1289},
  year={2018},
  publisher={ACS Publications}
}

@article{segler2018planning,
  title={Planning Chemical Syntheses with Deep Neural Networks and Symbolic {AI}},
  author={Segler, Marwin H S and Preuss, Mike and Waller, Mark P},
  journal={Nature},
  volume={555},
  number={7698},
  pages={604--610},
  year={2018},
  publisher={Nature Publishing Group}
}

@phdthesis{lowe2012extraction,
  title={Extraction of Chemical Structures and Reactions from the Literature},
  author={Lowe, Daniel Mark},
  school={University of Cambridge},
  year={2012},
  doi={10.17863/CAM.16293}
}

@article{burges2010ranknet,
  title={From {RankNet} to {LambdaRank} to {LambdaMART}: An Overview},
  author={Burges, Christopher JC},
  journal={Learning},
  volume={11},
  pages={23--581},
  year={2010}
}

@inproceedings{xiong2020layer,
  title={On Layer Normalization in the Transformer Architecture},
  author={Xiong, Ruibin and Yang, Yunchang and He, Di and Zheng, Kai and Zheng, Shuxin and Xing, Chen and Zhang, Huishuai and Lan, Yanyan and Wang, Liwei and Liu, Tie-Yan},
  booktitle={International Conference on Machine Learning (ICML)},
  pages={10524--10533},
  year={2020},
  organization={PMLR}
}

@article{zhong2022root,
  title={Root-aligned {SMILES}: A Tight Representation for Chemical Reaction Prediction},
  author={Zhong, Zipeng and Song, Jie and Feng, Zunlei and Liu, Tiantao and Jia, Lingxiang and Yao, Shaolun and Wu, Min and Liu, Tingjun and Song, Mingli},
  journal={Chemical Science},
  volume={13},
  number={31},
  pages={9023--9034},
  year={2022},
  publisher={Royal Society of Chemistry},
  doi={10.1039/D2SC02763A}
}

@inproceedings{press2017using,
  title={Using the Output Embedding to Improve Language Models},
  author={Press, Ofir and Wolf, Lior},
  booktitle={Proceedings of the 15th Conference of the European Chapter of the Association for Computational Linguistics (EACL)},
  pages={157--163},
  year={2017}
}

@article{han2024editretro,
  title={Retrosynthesis prediction with an iterative string editing model},
  author={Han, Yuqiang and Xu, Xiaoyang and Hsieh, Chang-Yu and Ding, Keyan and Xu, Hongxia and Xu, Renjun and Hou, Tingjun and Zhang, Qiang and Chen, Huajun},
  journal={Nature Communications},
  volume={15},
  number={1},
  pages={6404},
  year={2024},
  doi={10.1038/s41467-024-50617-1}
}

@article{zhong2023graph2edits,
  title={Retrosynthesis prediction using an end-to-end graph generative architecture for molecular graph editing},
  author={Zhong, Weihe and Yang, Ziduo and Chen, Calvin Yu-Chian},
  journal={Nature Communications},
  volume={14},
  number={1},
  pages={3009},
  year={2023},
  doi={10.1038/s41467-023-38851-5}
}

@article{chen2023g2retro,
  title={{G2Retro} as a two-step graph generative models for retrosynthesis prediction},
  author={Chen, Ziqi and Ayinde, Oludare R and Fuchs, Joseph R and Sun, Xia Ning and Ning, Xia},
  journal={Communications Chemistry},
  volume={6},
  number={1},
  pages={102},
  year={2023},
  doi={10.1038/s42004-023-00897-3}
}
\bibliographystyle{icml2026}

\clearpage
\onecolumn
\appendix
\section{Dataset and preprocessing details}
\label{app:data_details}

\begin{table}[H]
\centering
\caption{Dataset and generator-preprocessing statistics used in the draft. Augmentation is applied only to the training split.}
\small
\begin{tabular}{lcc}
\toprule
Statistic & Value & Note \\
\midrule
Train / val / test reactions & 40,008 / 5,000 / 5,007 & Standard \uspto\ split \\
Augmented training pairs & 800,160 & 20$\times$ offline augmentation \\
R-SMILES / random ratio & 16 / 4 & Per training reaction \\
Vocabulary size & 81 & Train-split only \\
Mean source / target length & 45.8 / 50.7 & Sampled 5K training pairs \\
1 / 2 / 3 precursor fragments & 29.1 / 70.7 / 0.2\% & Training reactions \\
\bottomrule
\end{tabular}
\end{table}

\section{Generator training and inference configuration}
\label{app:gen_config}
\begin{table}[H]
\centering
\caption{Generator configuration summary for the reproducible Baseline V2 generator used in this draft.}
\small
\begin{tabular}{lcc}
\toprule
Parameter & Value & Note \\
\midrule
Encoder / decoder layers & 6 / 6 & Transformer \\
$d_{\mathrm{model}}$ / heads / FFN & 512 / 8 / 2048 & Baseline V2 \\
Dropout & 0.1 & Training and decoding stack \\
Optimizer & Adam & $\beta_1=0.9$, $\beta_2=0.998$, $\epsilon=10^{-9}$ \\
Learning-rate schedule & Noam & factor 2.0, warmup 8,000 \\
Max tokens / accum steps & 16,384 / 2 & Dynamic token batching \\
Mixed precision / EMA & AMP fp16 / 0.999 & EMA checkpoint used for inference \\
Best checkpoint step & 20,000 & Early-stopped training \\
Beam search & max length 200 & Length norm $\alpha=0.6$ \\
Canonical test-time inference & Yes & Main table generator numbers \\
TTA extension & Available & Random product SMILES + aggregation \\
\bottomrule
\end{tabular}
\end{table}

\section{Generator ablation}
\label{app:generator_ablation}
\begin{table}[H]
\centering
\caption{Cumulative generator ablation on a 15K-reaction subset with 3K test reactions.}
\small
\begin{tabular}{llcc}
\toprule
Version & Added technique & Top-1 & Delta \\
\midrule
V1 & Baseline, 5x random augmentation & 34.97 & {--} \\
V6 & Larger Transformer & 36.60 & +1.63 \\
V7 & Hybrid R-SMILES & 45.83 & +9.23 \\
V8 & Pre-LN, EMA, atom-balance loss & 47.37 & +1.54 \\
V8-SWA & Stochastic weight averaging & 48.33 & +0.96 \\
V9 & Label smoothing and weight decay & 45.17 & -3.16 \\
\bottomrule
\end{tabular}
\end{table}

\section{Ranker protocol}
\label{app:ranker_protocol}
\begin{table}[H]
\centering
\caption{Compact summary of the LambdaMART reranking protocol.}
\small
\begin{tabular}{p{0.24\textwidth}p{0.68\textwidth}}
\toprule
Item & Setting \\
\midrule
Objective & XGBoost LambdaMART with \texttt{rank:ndcg}, optimized for NDCG@10 \\
Training groups & Candidates grouped by product, ranking never compares candidates across products \\
Labels & Graded relevance, exact precursor-set matches receive highest gain and partial fragment overlap receives weaker positive labels \\
Feature blocks & Upstream proposal score, structural descriptors, reaction-template descriptors, optional DFT descriptors \\
Template statistics & Frequency-style template features fit on the training split and frozen before validation/test scoring \\
Candidate pools & Train: 39,736 products, 29.9 candidates/product; val: 4,993 products, 30.1 candidates/product; test: 5,007 products, 111.4 candidates/product \\
Proposal coverage & Ground truth present in 96.3\% of train pools, 91.0\% of val pools, and 92.9\% of test pools \\
Observed signal & Upstream score is dominant, template-frequency features are next most useful, DFT features are secondary in the current runs \\
\bottomrule
\end{tabular}
\end{table}

\section{Ranker tuning summary}
\label{app:ranker_tuning}
\begin{table}[H]
\centering
\caption{Summary of the tuned LambdaMART configuration and sweep outcome.}
\small
\begin{tabular}{lcc}
\toprule
Parameter & Value & Note \\
\midrule
Objective / eval metric & \texttt{rank:ndcg} / NDCG@10 & Listwise ranking objective \\
Learning rate & 0.01 & Tuned from 0.05 \\
Max depth & 12 & Tuned from 5 \\
Min child weight & 6 & Tuned from 10 \\
Subsample / colsample & 0.8 / 0.4 & Tree regularization \\
Gamma / max delta step & 1.3 / 1 & Tuned stabilizers \\
Tree method & \texttt{hist} & CPU-friendly training \\
Rounds / early stopping & 5000 / 200 & Best round at 1121 \\
Sweep summary & 100 trials & Best val NDCG@10 improved from 0.7822 to 0.7912 \\
\bottomrule
\end{tabular}
\end{table}

\section{V1 ranker feature ablations}
\label{app:ranker_variants}
\begin{table}[H]
\centering
\caption{Feature ablations on the merged candidate-pool benchmark using the V1 proposal pool. Upstream proposal score is included in every row; only the additional feature blocks are listed. The spread across feature sets is below one percentage point on top-1, which motivated training the V2 reranker on the structural feature set only.}
\small
\begin{tabular}{lccccc}
\toprule
Feature set & Top-1 & Top-3 & Top-5 & Top-10 & MRR \\
\midrule
template & 58.07 & 81.50 & 87.35 & 91.16 & 0.7079 \\
structural + dft + template & 58.58 & 79.73 & 83.46 & 87.48 & 0.7011 \\
template + rc & 58.33 & 81.97 & 87.99 & 92.21 & 0.7123 \\
structural + dft + template + rc & 58.56 & 82.10 & 88.83 & 92.04 & \textbf{0.7168} \\
structural + DFT, no template & \textbf{58.76} & \textbf{82.34} & 87.97 & 91.60 & 0.7155 \\
A: structural + template & 58.56 & 81.73 & 87.99 & 91.52 & 0.7123 \\
B: structural + DFT + template & 58.56 & 81.45 & 86.94 & 90.90 & 0.7105 \\
C: DFT + template & 58.50 & 81.73 & 87.11 & 90.32 & 0.7094 \\
\bottomrule
\end{tabular}
\end{table}

\end{document}